\documentclass[letterpaper]{article} 
\usepackage{aaai25}  
\usepackage{times}  
\usepackage{helvet}  
\usepackage{courier}  
\usepackage[hyphens]{url}  
\usepackage{graphicx} 
\usepackage{natbib}  
\usepackage{caption} 
\usepackage{algorithm}
\usepackage{algorithmic}
\usepackage{amsmath}
\usepackage{amssymb}

\usepackage{newfloat}
\usepackage{listings}
\DeclareCaptionStyle{ruled}{labelfont=normalfont,labelsep=colon,strut=off} 
\floatstyle{ruled}
\newfloat{listing}{tb}{lst}{}
\floatname{listing}{Listing}
\title{Apollo-Forecast: Overcoming Aliasing and Inference Speed Challenges \\in Language Models for Time Series Forecasting}
\author {
    Tianyi Yin\textsuperscript{\rm 1,\rm 2},
    Jingwei Wang\textsuperscript{\rm 3}\thanks{Corresponding authors: Yunlong Ma, Jingwei Wang.},
    Yunlong Ma\textsuperscript{\rm 2}\footnotemark[1],
    Han Wang\textsuperscript{\rm 2},
    Chenze Wang\textsuperscript{\rm 2},\\
    Yukai Zhao\textsuperscript{\rm 2},
    Min Liu\textsuperscript{\rm 2},
    Weiming Shen\textsuperscript{\rm 4},
    Yufeng Chen\textsuperscript{\rm 5}
}
\affiliations {
    \textsuperscript{\rm 1}Shanghai Research Institute for Intelligent Autonomous Systems, Tongji University\\
    \textsuperscript{\rm 2}College of Electronics and Information Engineering, Tongji University\\
    \textsuperscript{\rm 3}Ant Group\\
    \textsuperscript{\rm 4}Huazhong University of Science and Technology\\
    \textsuperscript{\rm 5}Macau University of Science and Technology\\
    \{yin\_tianyi, evanma, 2210126, wangchenze, zhaoyukaijake, lmin\}@tongji.edu.cn, \\wangjingwei.wjw@antgroup.com, wshen@ieee.org, yfchen@must.edu.mo
}

\usepackage{bibentry}

\begin{document}

\maketitle

\begin{abstract}
Encoding time series into tokens and using language models for processing has been shown to substantially augment the models' ability to generalize to unseen tasks. 
However, existing language models for time series forecasting encounter several obstacles, including aliasing distortion and prolonged inference times, primarily due to the limitations of quantization processes and the computational demands of large models.
This paper introduces Apollo-Forecast, a novel framework that tackles these challenges with two key innovations: the Anti-Aliasing Quantization Module (AAQM) and the Race Decoding (RD) technique. AAQM adeptly encodes sequences into tokens while mitigating high-frequency noise in the original signals, thus enhancing both signal fidelity and overall quantization efficiency. RD employs a draft model to enable parallel processing and results integration, which markedly accelerates the inference speed for long-term predictions, particularly in large-scale models. 
Extensive experiments on various real-world datasets show that Apollo-Forecast outperforms state-of-the-art methods by 35.41\% and 18.99\% in WQL and MASE metrics, respectively, in zero-shot scenarios. Furthermore, our method achieves a 1.9X-2.7X acceleration in inference speed over baseline methods. 
\end{abstract}

%

\section{Introduction}

Time series forecasting plays a pivotal role in various research domains, including transportation \cite{afrin2022long}, energy \cite{cai2024msgnet}, and manufacturing \cite{wang2022few}. Early forecasting techniques predominantly relied on statistical methods such as ARIMA and machine learning models \cite{li2023self}. These approaches are often suitable for scenarios with limited observational data, offering a balanced performance. The emergence of deep learning techniques, such as NHITS \cite{challu2023nhits} and Dlinear \cite{Zeng2022AreTE}, coinciding with the expansion of diverse time series data sources, has significantly enhanced forecasting capabilities \cite{tang2022survey}. Nonetheless, these methods typically operate under a one-model-per-dataset framework and have yet to achieve a universally applicable forecasting model \cite{zhang2024large}.

The emergence of large language models (LLMs) with robust generalization capabilities has inspired the development of foundation models for time-series forecasting. Those studies can be categorized into two main fields: enhancing pre-trained LLMs with prompts and retraining LLMs for time series tasks \cite{zhang2024large}. These methods bridge the gap between specialized models and general approaches, eliminating the need for retraining from scratch for each task \cite{jin2024time}. Notably, the latter approach avoids errors introduced by the generic LLM vocabulary by encoding the time series into tokens, also known as quantization.

\begin{figure}[t]
\centering
\includegraphics[width=1.0\columnwidth]{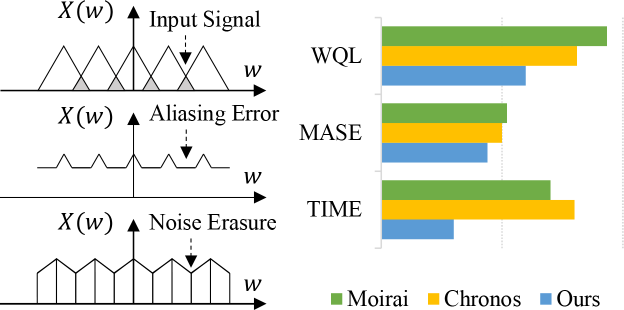}
\caption{\textbf{(Left)} Without a noise erasure mechanism, quantization causes aliasing distortion. With a noise erasure mechanism, high-frequency noise is removed, preserving low-frequency information. $X$ represents the frequency domain. \textbf{(Right)} Performance comparison of Moirai-L, Chronos-S, and Apollo-S on UCR dataset.}
\label{fig:alisaing}
\end{figure}

Despite the advantages of those foundation models, existing quantization methods may introduce high-frequency noise into the low-frequency band during tokenization, leading to additional distortion known as aliasing errors \cite{ma2024high}, as illustrated in Figure \ref{fig:alisaing}. Furthermore, due to the autoregressive mechanism and the large model size, the response time of these models is relatively slow, especially in long-horizon scenarios \cite{cleaveland2024conformal}. These issues limit the model's performance in real situations and increase the cost of utilization.

\begin{figure*}[t]
\centering
\includegraphics[width=2.0\columnwidth]{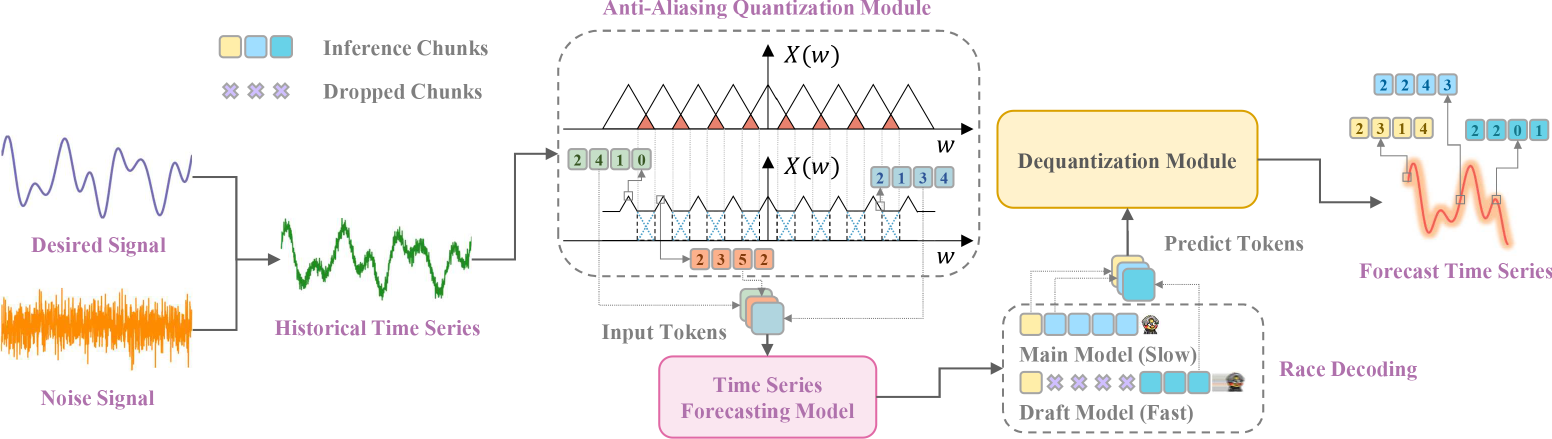}
\caption{The architecture of Apollo-Forecast. The time series first passes through the \textbf{AAQM} to be converted into tokens, and then the forecasting model with \textbf{Race Decoding} is used to predict the next value. $X$ represents the frequency domain.}
\label{fig:flowchart}
\end{figure*}

In this paper, we propose a novel Apollo-Forecast \footnote{Apollo is recognized as the god of truth and prophecy.} framework to address these challenges, incorporating two critical improvements: the Anti-Aliasing Quantization Module (AAQM) and Race Decoding (RD). AAQM introduces a Noise Erasure mechanism to eliminate interference components from the original signal, preventing information loss due to quantization with a fixed token length. The RD technique accelerates inference by employing a draft model (a faster model with reasonable accuracy) and a main model for parallel processing, directly merging the draft results with the main results after passing a Tolerance Check. We conducted extensive experiments across various datasets in multiple domains, demonstrating the generalization and speed enhancements of the proposed method.

Our contributions are summarized as follows:

\begin{enumerate}
    \item We present the Anti-Aliasing Quantization module that substantially reduces aliasing errors caused by the encoding process.
    \item We propose a novel decoding mechanism called Race Decoding to accelerate the inference speed of the foundation model without altering any structure.
    \item We demonstrate through extensive experiments that our method improves performance by 35.41\% in weighted quantization loss (WQL) and speed by 1.9X-2.7X compared to SOTA methods.
\end{enumerate}

\section{Related Work}
\subsection{Time Series Forecasting}

Time series forecasting involves predicting future trends based on historical data and is widely used in transportation \cite{deng2022multi}, manufacturing \cite{he2023learning}, energy \cite{zhang2023ctfnet}, and other fields. The methods for time series forecasting can be broadly categorized into three types: statistical models, machine learning models, and deep learning models \cite{chen2023long}.

\subsubsection{Statistical Models}
Early approaches primarily utilized approaches such as ARIMA and EMA Model \cite{ansari2024chronos}. These models rely on mathematical formulations to capture underlying patterns in the data. ARIMA models are particularly effective for univariate time series and can handle non-stationary data through differencing. EMA Model, based on exponential smoothing, capturing seasonality and trend components \cite{du2022bayesian}. While these models can produce reasonable outcomes from limited observations, they often require manual tuning and domain expertise.

\subsubsection{Machine Learning Predictors}
Machine learning introduced more sophisticated models like Support Vector Machines \cite{vukovic2022cds}, Random Forests, and Gradient Boosting Machines for time series forecasting \cite{li2023deep}. These models can capture complex patterns and interactions in the data without explicit mathematical formulations. Machine learning models often involve feature engineering to extract relevant features from raw time series data, enhancing predictive performance. However, they may struggle with capturing long-term dependencies and require careful hyperparameter tuning.

\subsubsection{Deep Learning Predictors}
The development of deep learning has led to new feature extractors such as RNN-based, MLP-based, CNN-based, and Transformer-based models. RNNs, including LSTM and GRU, are particularly effective for sequential data as they capture temporal dependencies \cite{li2023deep}. CNNs can extract local patterns and features from time series data \cite{he2022rel}. Transformer-based models, which have shown impressive performance in natural language processing, are being adapted for time series forecasting due to their ability to capture long-range dependencies and parallelize computations \cite{chowdhury2022tarnet}. Despite their outstanding performance, these deep learning methods are still limited to a one-model-per-dataset paradigm, making it challenging to extend a single model to datasets from different domains and failing to provide an integrated and general predictor.

\subsection{LLMs for Time Series}

Encoding time series as digits and processing them with LLMs offers a general approach without the need to train from scratch for different datasets \cite{gruver2023llmtime}. Current research bridges the gap between different modalities through three methods: prompts, multimodal alignment, and discretization. The prompt-based method uses natural language to describe the data, enabling the foundation model to develop analytical capabilities through in-context learning or fine-tuning \cite{jin2024position}. Multimodal alignment maps the feature representations of time series to the text space through contrastive learning, allowing time series data to be directly input into LLMs for prediction. Discretization converts observed data into special discrete identifiers (IDs) and then passes them to the foundation model, avoiding the shortcomings of traditional dictionaries in distinguishing numbers. However, these methods are limited by token length requirements during the conversion process, leading to aliasing distortion that affects encoding accuracy. Additionally, these models are often based on LLMs such as LLaMA and GPT-2 \cite{rasul2024lagllama, jin2023time}, which have slower inference speeds, impacting practical use. This study aims to address these issues.

\section{Methodology}
The architecture of the Apollo-Forecast framework is illustrated in Figure \ref{fig:flowchart}. It consists of three modules: the Anti-Aliasing Quantization Module (AAQM), the time series forecasting model (TSFM), and the Race Decoding (RD). Each component plays a distinct role within the framework, and their details will be introduced in the following.

\subsection{Anti-Aliasing Quantization Module}

When the token sequence length is fixed and cannot meet the requirements of the Nyquist sampling theorem, high-frequency noise will fold back into the low-frequency band, affecting the signal quality \cite{ma2024high}. To address this issue, we propose a novel method for time series tokenization called AAQM. This module comprises a noise erasure and a time series embedding module to achieve higher quality temporal signal encoding, as shown in Algorithm~\ref{alg:anti_aliasing_quantization}.

\subsubsection{Noise Erasure}

Specifically, in the Noise erasure (NE) mechanism, we employ a Butterworth filter to remove high-frequency noise from the raw signal. Let \( x[n] \) be the original discrete-time signal, and \( X[k] \) its Discrete Fourier Transform (DFT). The filter is designed to have a maximally flat frequency response in the passband. The transfer function \( H(z) \) of an \( n \)-th order Butterworth filter in the z-domain is given by:

\begin{equation}
    H(z) = \frac{1}{\sqrt{1 + \left(\frac{z}{\omega_c}\right)^{2n}}} ,
    \label{eq:butterworth}
\end{equation}
where \( \omega_c \) is the low cutoff frequency, and \( z = e^{j\omega} \) is the complex frequency variable. The filtered signal \( y[n] \) is obtained by applying the inverse DFT ($\mathcal{F}^{-1}$) to the product of \( X[k] \) and \( H(e^{j\omega}) \):

\begin{equation}
    y[n] = \mathcal{F}^{-1}\left\{ X[k] \cdot H(e^{j\omega}) \right\} .
\end{equation}

Assume the input signal \( x[n] \) is composed of the desired signal \( s[n] \) and high-frequency noise \( e[n] \), i.e., \( x[n] = s[n] + e[n] \). The power spectral densities of signal and noise are \( S_[k]\) and \( S_e[k]\), respectively. Before applying the NE mechanism, the signal-to-noise ratio (SNR) can be expressed as:

\begin{equation}
    \text{SNR} = \frac{\sum_{k=0}^{N-1} |S[k]|^2}{\sum_{k=0}^{N-1} S_e[k]} ,
    \label{eq:snr_before}
\end{equation}

After applying the NE operation, the power of the filtered signal \( P_{s}' \) and noise \( P_{n}' \) are:

\begin{equation}
    P_{s}' = \sum_{k=0}^{\omega_c} |S[k]|^2 ,
\end{equation}

\begin{equation}
    P_{n}' = \sum_{k=0}^{\omega_c} S_e[k] |H(e^{j\omega})|^2 ,
\end{equation}

Thus, the signal-to-noise ratio after filtering is:

\begin{equation}
    \text{SNR}' = \frac{P_{s}'}{P_{n}'} = \frac{\sum_{k=0}^{\omega_c} |S[k]|^2}{\sum_{k=0}^{\omega_c} S_e[k] |H(e^{j\omega})|^2} ,
    \label{eq:snr_after}
\end{equation}

Most of the real signals in nature usually have the following properties:

\begin{itemize}
\item The desired signal \( s[n] \) has most of its energy concentrated in the low-frequency range \( [0, \omega_c] \) .
\item The high-frequency noise \( e[n] \) has significant energy in the high-frequency range \( [\omega_c, N-1] \) .
\end{itemize}

Based on these conditions, the NE mechanism effectively attenuates the high-frequency noise while preserving the low-frequency components of the input. Therefore, the power of the desired signal \( P_{s}' \) remains largely unchanged, while the power of the noise \( P_{n}' \) is significantly reduced.

Mathematically, this can be expressed as:

\begin{equation}
    \sum_{k=0}^{\omega_c} |S[k]|^2 \approx \sum_{k=0}^{N-1} |S[k]|^2 ,
\end{equation}

\begin{equation}
    \sum_{k=0}^{\omega_c} S_e[k] |H(e^{j\omega})|^2 \ll \sum_{k=0}^{N-1} S_e[k] ,
\end{equation}

Thus, we have:

\begin{equation}
    \frac{\sum_{k=0}^{\omega_c} |S[k]|^2}{\sum_{k=0}^{\omega_c} S_e[k] |H(e^{j\omega})|^2} > \frac{\sum_{k=0}^{N-1} |S[k]|^2}{\sum_{k=0}^{N-1} S_e[k]} .
\end{equation}

The proof demonstrates that the proposed method effectively reduces high-frequency noise, increasing the signal-to-noise ratio and reducing aliasing distortion. Compared to other filtering methods, our approach does not affect the phase, eliminating the need for phase compensation and thereby reducing computational load.

\begin{algorithm}[tb]
\caption{Anti-Aliasing Quantization}
\label{alg:anti_aliasing_quantization}
\textbf{Input}: $x[n]$, $\omega_c$, $f_s$, $n$, $Q$\\
\textbf{Output}: Token sequence $\mathcal{T}$
\begin{algorithmic}[1]
\STATE $f_N \gets \frac{f_s}{2}$
\STATE $\omega_N \gets \frac{\omega_c}{f_N}$
\STATE Compute filter coefficients $b, a \gets \text{butter}(n, \omega_N, \text{low})$
\STATE Apply zero-phase filtering $y[n] \gets \text{filtfilt}(b, a, x[n])$
\STATE Normalize $y_{norm}[n] \gets \frac{y[n] - \min(y[n])}{\max(y[n]) - \min(y[n])}$
\STATE Quantize $y_{q}[n] \gets \left\lfloor y_{norm}[n] \times Q \right\rfloor / Q$
\STATE Tokenize $\mathcal{T} \gets \left\{ t_i \mid t_i = \mathcal{E}(y_{q}[i]), \forall i \in [1, N] \right\}$
\STATE \textbf{return} $\mathcal{T}$
\end{algorithmic}
\end{algorithm}

\subsubsection{Time Series Embedding Module}

In order to use LLM for prediction, we convert the obtained high-quality series signal into a discrete token representation. In this process, the complete time series signal can be considered as \( \mathbf{x} = \{x_1, x_2, \ldots, x_N\} \), where \( N \) is the length of the sequence. The signal is first normalized and quantized as follows:

\begin{equation}
    x_{norm} = \frac{\mathbf{x} - \min(\mathbf{x})}{\max(\mathbf{x}) - \min(\mathbf{x})},
    \label{eq:shaping1}
\end{equation}

\begin{equation}
    x_{q} = \left\lfloor x_{norm} \times Q \right\rfloor / Q,
    \label{eq:shaping2}
\end{equation}
where \( Q = 10^4 \) is the quantization factor, and \( \left\lfloor \cdot \right\rfloor \) denotes the floor function, which rounds down to the nearest integer. The resulting quantized sequence is \( \mathbf{x}_{q} = \{x_{q,1}, x_{q,2}, \ldots, x_{q,N}\} \). The quantized values are retained to four decimal places, ranging from 0.0000 to 1.0000, with a theoretical quantization error within the interval \([- \frac{1}{2Q}, \frac{1}{2Q}]\).

The tokenization process involves mapping the normalized and quantized signal \( \mathbf{x}_{q} \) to a discrete set of tokens. The set of tokens is denoted as \( \mathcal{T} \), and the embedding module as \( \mathcal{E} \). The tokenization can be expressed as:

\begin{equation}
    \mathcal{T}(\mathbf{x}_{q}) = \left\{ t_i \mid t_i = \mathcal{E}(x_{q,i}), \forall i \in [1, N] \right\},
\end{equation}

Where \( N \) is the length of the time series. The embedding function \( \mathcal{E} \) maps each quantized value to a token in the vocabulary. The resulting token sequence \( \mathcal{T}(\mathbf{x}_{q}) \) is then used as input for the LLM.

\subsection{Time Series Forecasting Model}


By converting time series into tokens, we can predict the next time frame similarly to the seq2seq tasks in natural language processing. In this process, the TSFM can be viewed as a transformer that converts historical observations into forecast sequences, as illustrated by the following formula:

\begin{equation}
    \hat{y}_{i+h} = f(t_{i}, t_{i-1}, \ldots, t_{i-n}; \theta),
\end{equation}

Here, \(\hat{y}_{i+h}\) represents the predicted value at time \(i+h\), where \(h\) is the prediction horizon. The function \(f\) takes as input the historical token sequences \(t_{i}, t_{i-1}, \ldots, t_{i-n}\), which are the observed values at times \(i, i-1, \ldots, i-n\) respectively. The parameter \(\theta\) denotes the model parameters.

Our approach involves predicting frames with a fixed horizon based on the input historical data and comparing them with the ground truth. The training loss is measured using the cross-entropy function:

\begin{equation}
    \mathcal{L} = -\sum_{i=1}^{N} y_i \log(\hat{y}_i),
\end{equation}

The data for training LLM includes various domains with different time granularities, such as finance, electricity, and transportation. This diversity enhances the model's generalization capability in zero-shot scenarios \cite{das2023decoder}.

\subsection{Race Decoding}

Despite the seq2seq model performing well, its inference speed significantly slows down when dealing with long-horizon prediction scenarios. Existing speculative decoding methods \cite{leviathan2023fast} can accelerate large language model predictions at the token level, but they result in substantial computational waste for time series prediction tasks. Therefore, we propose the RD mechanism to accelerate the inference speed of TSFM at the chunk level while maintaining accuracy, as shown in Algorithm~\ref{alg:RD_algorithm}.

\subsubsection{Draft Inference}

Models with fewer parameters generally have faster inference speeds under the same computational precision. In our method, when using a large model as the primary predictor for time series, a smaller, auxiliary model is automatically assigned as the draft predictor. Both predictors iteratively forecast the time series frames in parallel. Given the prediction horizon $H$, the inference time $T$ for the primary and draft predictors can be defined as functions of $H$. As $H \to \infty$, the speed advantage of the draft predictor becomes more apparent, i.e., $T_{\text{main}}/{T_{\text{draft}}} \gg 1$. This implies that the draft predictor allows for quicker complete results as the horizon increases.

Ideally, during training, both the draft and primary predictors learn to approximate the same probability distribution $P(y|x)$ with different precisions. Let $\mathcal{D}$ represent the data distribution, and $\mathcal{L}$ be the loss function. The training objective for both predictors can be expressed as:

\begin{equation}
\theta_{\text{main}}^* = \arg\min_{\theta_{\text{main}}} \mathbb{E}_{(x,y) \sim \mathcal{D}} [\mathcal{L}(P_{\text{main}}(y|x; \theta_{\text{main}}), y)],
\end{equation}

\begin{equation}
\theta_{\text{draft}}^* = \arg\min_{\theta_{\text{draft}}} \mathbb{E}_{(x,y) \sim \mathcal{D}} [\mathcal{L}(P_{\text{draft}}(y|x; \theta_{\text{draft}}), y)].
\end{equation}

Here, $\theta_{\text{main}}$ and $\theta_{\text{draft}}$ represent the parameters of the primary and draft predictors, respectively. The notation $\theta_{\text{main}}^*$ and $\theta_{\text{draft}}^*$ denote the optimal parameters that minimize the expected loss $\mathcal{L}$ over the data distribution $\mathcal{D}$. With sufficient training, both predictors should ideally approximate the true distribution $P(y|x)$, with the errors due to model complexity denoted by $\epsilon_{\text{main}}$ and $\epsilon_{\text{draft}}$:

\begin{equation}
P_{\text{main}}(y|x) \approx P(y|x) + \epsilon_{\text{main}},
\end{equation}

\begin{equation}
P_{\text{draft}}(y|x) \approx P(y|x) + \epsilon_{\text{draft}}.
\end{equation}

When both predictors have sufficient representational capability to fit the current data distribution, the errors due to model complexity $\epsilon_{\text{main}}$ and $\epsilon_{\text{draft}}$ also tend to zero. Thus, we can approximate:

\begin{equation}
\lim_{\epsilon_{\text{main}}, \epsilon_{\text{draft}} \to 0} P_{\text{main}}(y|x) \approx P_{\text{draft}}(y|x).
\end{equation}

However, due to model randomness and distribution drift, the actual output distributions of the two predictors may differ. Therefore, after the draft predictor completes inference, an additional tolerance check step is required.

\begin{algorithm}[tb]
\caption{Race Decoding Algorithm}
\label{alg:RD_algorithm}
\textbf{Input}: $x$, $H$\\
\textbf{Output}: $y$
\begin{algorithmic}[1] 
\STATE Initialize $\theta_{\text{main}}$ and $\theta_{\text{draft}}$
\STATE $t=0$
\WHILE{$t < H$}
    \STATE \textbf{Draft Inference:}
    \STATE $P_{\text{draft}}(y|x; \theta_{\text{draft}})$
    \STATE $P_{\text{main}}(y|x; \theta_{\text{main}})$
    \STATE \textbf{Tolerance Check:}
    \STATE $\Delta P_t = \|P_{\text{main}}(y_{t_1:t_k}|x) - P_{\text{draft}}(y_{t_1:t_k}|x)\|$
    \IF {$\Delta P_t < \gamma$}
        \STATE \textbf{Result Concatenation:}
        \STATE Concatenate $P_{\text{main}}(y_{t_1:t_k}|x)$ with $P_{\text{draft}}(y_{t_{k+1}:t_H}|x)$
        \STATE \textbf{return} concatenated result
    \ELSE
        \STATE Wait for $\theta_{\text{main}}$ to complete inference
        \STATE \textbf{return} $P_{\text{main}}(y|x; \theta_{\text{main}})$
    \ENDIF
    \STATE $t = t + 1$
\ENDWHILE
\end{algorithmic}
\end{algorithm}

\subsubsection{Tolerance Check}

Since the output probability distribution of the same TSFM under fixed temperature parameters is not prone to sudden changes, we can compare the inference results of the primary predictor with the corresponding frames of the draft predictor. When the time difference between the inferences of the two predictors is small, we can use the distribution $P$ composed of partial frames as a proxy for the full sequence. Let $t_1:t_k$ denote the frames predicted by the primary predictor and $t_{k+1}:t_H$ denote the frames predicted by the draft predictor. Then:

\begin{equation}
\Delta P_t = \|P_{\text{main}}(y_{t_1:t_k}|x) - P_{\text{draft}}(y_{t_1:t_k}|x)\|,
\end{equation}

As $\Delta P_t$ approaches zero, the real difference between the main and draft predictors' output distributions can be expressed as:

\begin{equation}
\lim_{\Delta P_t \to 0} \|P_{\text{main}}(y|x) - P_{\text{draft}}(y|x)\| \approx 0,
\end{equation}

When the two proxies are consistent in value, it can be approximated that the two outputs follow the same distribution. However, in time series prediction, the result fluctuation range is large, and strictly limiting consistency may render the acceleration mechanism ineffective. Therefore, we introduce an error factor $\gamma$ to save computational cost while enhancing generalization, as shown in the following:

\begin{equation}
\|P_{\text{main}}(y_t|x) - P_{\text{draft}}(y_t|x)\| < \gamma.
\end{equation}

This mechanism also achieves an ensemble learning effect, reducing the possibility of overfitting in large models.

\subsubsection{Result Concatenation}

After passing the tolerance check, our method concatenates the existing inference results of the primary predictor with the remaining results of the draft predictor as the final result to further reduce errors. To compare the errors, let $\delta_{\text{draft}}$ be the error when using only the draft predictor, and $\eta_{\text{concat}}$ be the error when concatenating the primary predictor's results with the draft predictor's results, as shown in the following equation:

\begin{equation}
\eta_{\text{concat}} = \sum_{t=1}^{k} \delta_{\text{main}, t} + \sum_{t=k+1}^{H} \delta_{\text{draft}, t}.
\end{equation}

Assume that the primary predictor completes $k$ frames, and the draft predictor completes the remaining $H-k$ frames. Since the error for each frame predicted by the primary predictor ($\delta_{\text{main}, t}$) is typically smaller than the error for each frame predicted by the draft predictor ($\delta_{\text{draft}, t}$), the total error $\eta_{\text{concat}}$ when concatenating the results is smaller than the total error $\delta_{\text{draft}}$ when using only the draft predictor.

Given that the error for each frame predicted by the draft predictor is greater than or equal to the error for each frame predicted by the primary predictor, i.e., $\delta_{\text{draft}, t} \geq \delta_{\text{main}, t}$, it follows that the total error $\eta_{\text{concat}}$ when concatenating the results is indeed smaller than the total error $\delta_{\text{draft}}$ when using only the draft predictor.

If the draft results do not pass the tolerance check, we continue to wait for the primary predictor to complete the inference and use it as the complete return result. The theoretical shortest time for the RD mechanism is the sum of the draft predictor inference time and the tolerance comparison time (extremely fast), and the longest time is the primary predictor inference time, mathematically expressed as:

\begin{equation}
T_{\text{RD}} = \min(T_{\text{draft}} + T_{\text{tolerance}}, T_{\text{main}}).
\end{equation}

This ensures that the Race Decoding mechanism effectively accelerates the inference process while maintaining the accuracy of the predictions.

\section{Experiments}

\subsection{Baselines and Experimental Settings}
To evaluate the performance of the proposed method on real datasets, we selected over ten SOTA algorithms covering LLM-based (Chronos, Moirai), RNN-based (LSTM), MLP-based (N-HiTS, N-BEATS, DLinear), Transformer-based (TFT), and statistical models (AutoARIMA, AutoETS) as baselines for comparison experiments. Regarding datasets, we conducted extensive experiments on different datasets to verify the generalization of the proposed method under zero-shot conditions. In the experiments, we used five pre-trained Chronos-T5 models \cite{ansari2024chronos} of different sizes as TSFM, namely Tiny (T), Mini (M), Small (S), Base (B), and Large (L), with a default horizon of 64. Meanwhile, the tiny model with AAQM is used as a draft model with a precision of float32, labeled Apollo-T*.

\begin{figure*}[t]
\centering
\includegraphics[width=2.1\columnwidth]{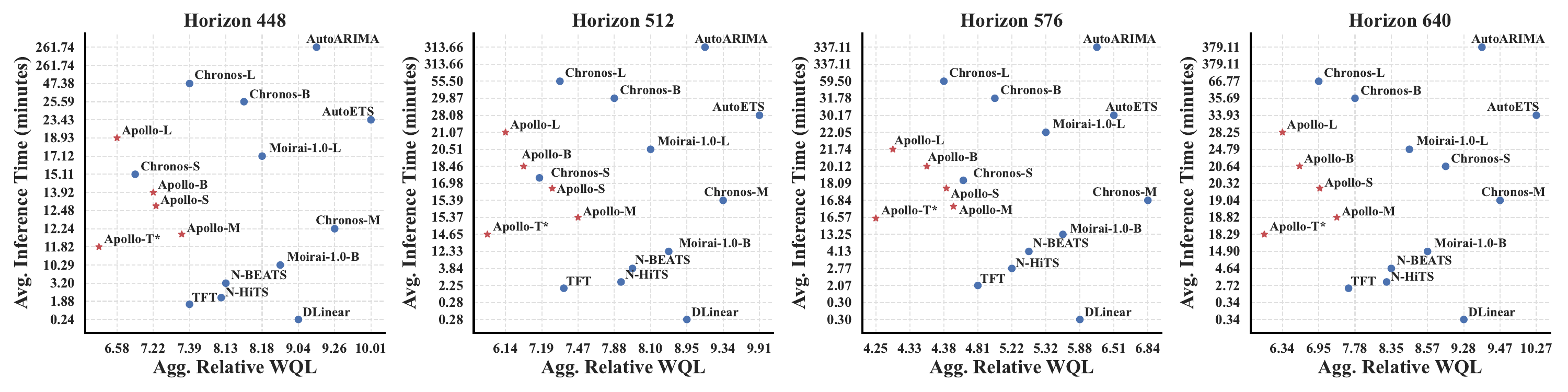}
\caption{The Agg. Relative WQL of our Apollo-Forecast approach and other baseline models on the UCR dataset.}
\label{fig:ucrper}
\end{figure*}

\begin{figure*}[t]
\centering
\includegraphics[width=2.1\columnwidth]{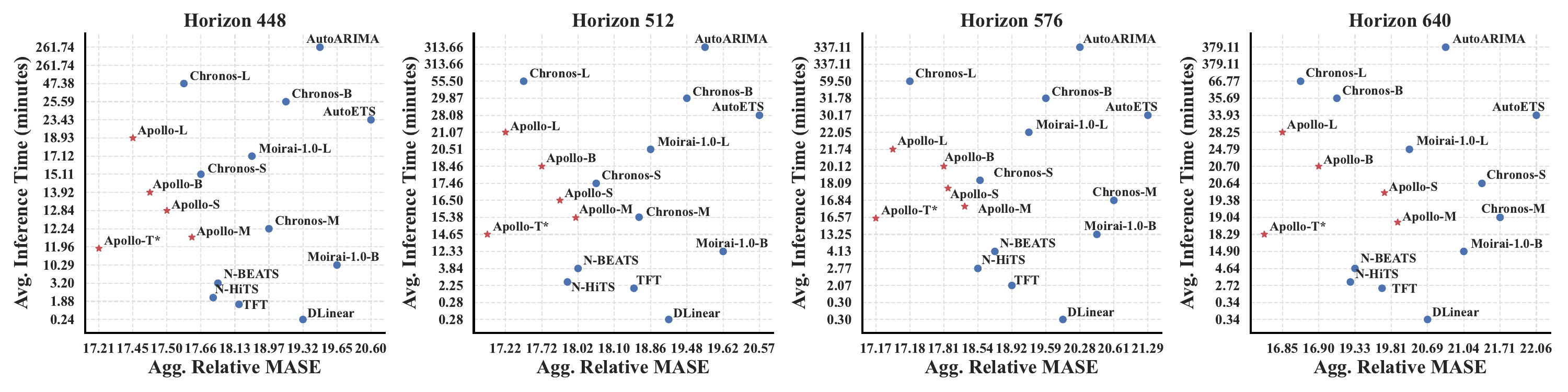}
\caption{The Agg. Relative MASE of our Apollo-Forecast approach and other baseline models on the UCR dataset.}
\label{fig:ucrtime}
\end{figure*}

\subsection{Main Results}

\subsubsection{UCR Dataset}
In this experiment, we selected 114 available records from the UCR dataset for the experiment \cite{dau2019ucr}, covering multiple fields such as health, energy, and traffic. The prediction horizon ranges from 448 to 640, and the inference precision uses bfloat16. Regarding the hyperparameters of AAQM, we set the cutoff to 10 and the order to 5. During the evaluation phase, we separately counted the impact of AAQM and RD on Avg. Inference Time, Agg. Relative WQL, and Agg. Relative MASE, with other baseline methods using the same settings \cite{ansari2024chronos}.

As illustrated in Figure \ref{fig:ucrper} and \ref{fig:ucrtime}, the accuracy enhancement provided by Apollo-Forecast becomes increasingly evident with longer prediction horizons (Tiny models are excluded due to negligible effects). In the best cases, it achieves a reduction in Agg. Relative WQL by up to 35.41\% and in Agg. Relative MASE by up to 17.01\%. In terms of inference speed, RD significantly improves performance by increasing the average inference time by as much as 2.7X. These improvements are particularly pronounced in scenarios involving long horizons and large models.

\subsubsection{Public Dataset} 

In this experiment, we selected seven types of records as input: \textit{ETTh1}, \textit{ETTh2}, \textit{ETTm1}, \textit{ETTm2}, \textit{Weather}, \textit{ECL}, and \textit{TrafficL} \cite{zhou2021informer}. The prediction horizon range is equal to the length of the test set, and the inference precision uses bfloat16. Regarding the hyperparameters of AAQM, we set the cutoff to 50 and the order to 5. During the evaluation phase, we measured the impact of AAQM and RD on inference time (minutes), WQL, and MASE with other baseline methods using the same settings.

\begin{table*}[ht]
    \centering
    \small
    \begin{tabular}{c|cc|cc|cc|cc|cc|ccccc}
        \hline
        Methods & \multicolumn{2}{c|}{Apollo-S} & \multicolumn{2}{c|}{Apollo-B} & \multicolumn{2}{c|}{Apollo-L} & \multicolumn{2}{c|}{Chronos-S} & \multicolumn{2}{c|}{Chronos-B} & \multicolumn{2}{c}{Chronos-L} \\
        \hline
        Metric & WQL & MASE & WQL & MASE & WQL & MASE & WQL & MASE & WQL & MASE & WQL & MASE \\
        \hline
        ETTh1 & 2.135 & 4.120 & \textbf{0.265} & 3.869 & 0.267 & 4.049 & 2.309 & 4.421 & 0.307 & \textbf{3.849} & 0.339 & 4.276 \\
        ETTh2 & 2.124 & 4.107 & \textbf{0.255} & \textbf{3.704} & 0.286 & 4.100 & 2.321 & 4.414 & 0.287 & 3.883 & 0.332 & 4.263 \\
        ETTm1 & 2.126 & 4.120 & 0.270 & \textbf{3.702} & \textbf{0.261} & 4.149 & 2.327 & 4.418 & 0.292 & 3.926 & 0.336 & 4.327 \\
        ETTm2 & 2.135 & 4.102 & \textbf{0.257} & \textbf{3.698} & 0.291 & 4.081 & 2.323 & 4.446 & 0.287 & 4.073 & 0.334 & 4.138 \\
        Weather & 2.135 & 4.129 & \textbf{0.235} & \textbf{3.655} & 0.292 & 3.697 & 2.298 & 4.390 & 0.343 & 3.907 & 0.350 & 3.967 \\
        ECL & 2.163 & 4.105 & \textbf{0.217} & \textbf{3.741} & 0.255 & 3.785 & 2.306 & 4.442 & 0.292 & 4.163 & 0.284 & 4.672 \\
        TrafficL & 2.092 & 4.094 & \textbf{0.252} & 4.022 & 0.308 & 4.279 & 2.286 & 4.411 & 0.292 & \textbf{3.920} & 0.335 & 4.248 \\
        \hline
    \end{tabular}
    \caption{The performance of Apollo-Forecast and Chronos on public datasets.}
    \label{tab:public}
\end{table*}

\begin{table*}[htbp]
    \centering
    \small
    \begin{tabular}{c|ccc|cccc}
        \hline
        Dataset & Apollo-S & Apollo-B & Apollo-L & Chronos-S & Chronos-B & Chronos-L \\
        \hline
        ETTh1 & 8.479 ± 0.122 & 9.866 ± 0.181 & 19.904 ± 0.243 & 10.758 ± 0.139 & 19.252 ± 0.193 & 36.488 ± 0.253 \\
        ETTh2 & 8.371 ± 0.117 & 9.853 ± 0.170 & 20.093 ± 0.237 & 10.569 ± 0.127 & 19.151 ± 0.185 & 36.429 ± 0.262 \\
        ETTm1 & 8.489 ± 0.123 & 9.810 ± 0.174 & 20.021 ± 0.234 & 10.820 ± 0.134 & 19.211 ± 0.195 & 36.477 ± 0.245 \\
        ETTm2 & 8.235 ± 0.115 & 9.734 ± 0.188 & 20.019 ± 0.224 & 10.819 ± 0.121 & 19.129 ± 0.184 & 36.488 ± 0.255 \\
        Weather & 8.249 ± 0.131 & 9.866 ± 0.162 & 19.859 ± 0.250 & 10.501 ± 0.140 & 19.061 ± 0.171 & 36.362 ± 0.261 \\
        ECL & 8.469 ± 0.105 & 9.965 ± 0.193 & 20.098 ± 0.245 & 10.801 ± 0.115 & 19.377 ± 0.185 & 36.621 ± 0.277 \\
        TrafficL & 8.297 ± 0.136 & 9.996 ± 0.164 & 19.935 ± 0.223 & 10.702 ± 0.142 & 19.219 ± 0.170 & 36.495 ± 0.231 \\
        \hline
    \end{tabular}
    \caption{The inference time (minutes) of Apollo-Forecast and Chronos on public datasets.}
    \label{tab:public_times}
\end{table*}

The results are shown in Table \ref{tab:public} and \ref{tab:public_times}. The accuracy improvement brought by Apollo-Forecast in zero-shot scenarios is very significant, with the highest reductions in WQL and MASE being 31.44\% and 18.99\%, respectively, compared to the baseline. In terms of inference speed, our approach can accelerate average inference time by up to 1.9X.

\subsubsection{LBS Dataset}

In this experiment, we utilized crowd flow data from a specific area in Shanghai, spanning from 1/1/2023 to 7/1/2024—the data was recorded at 5-minute intervals. We fixed the horizon at 100 and used bfloat16 for inference. Both Apollo and Chronos models were set to a small size. For AAQM, we chose a cutoff of 30 and an order of 3. Since the RD mechanism was not employed in this experiment, we focused solely on the impact of AAQM on metrics, applying the same settings to other methods.

\begin{table}[ht]
    \centering
    \small
    \begin{tabular}{c|c|c|c|c}
        \hline
        Methods & MASE & MAE & MSE & WQL \\
        \hline
        \textbf{Apollo-S} & \textbf{0.24} & \textbf{63.44} & \textbf{0.10} & \textbf{0.10} \\
        Chronos-S & 0.28 & 76.37 & 0.30 & 0.30 \\
        N-HiTS & 0.37 & 99.48 & 0.72 & 0.60 \\
        N-BEATS & 0.40 & 107.37 & 0.99 & 0.93 \\
        TFT & 0.39 & 105.25 & 0.91 & 0.82 \\
        LSTM & 0.42 & 113.24 & 1.00 & 1.00 \\
        \hline
    \end{tabular}
    \caption{Performance of different algorithms on LBS. The values of MSE and WQL are scaled.}
    \label{tab:lbs_per}
\end{table}

\begin{figure}[ht]
\centering
\includegraphics[width=0.8\columnwidth]{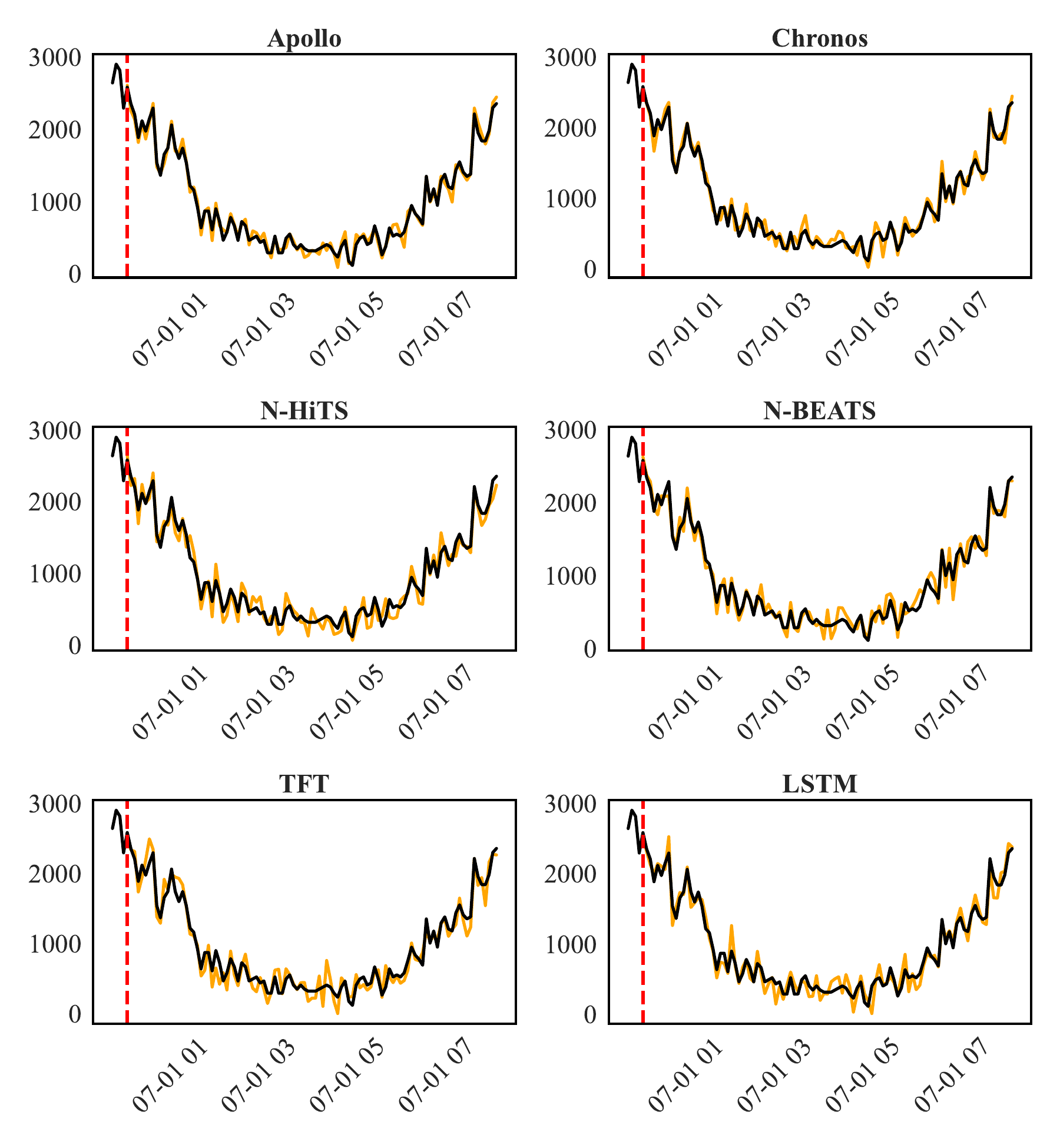}
\caption{The results of our method and other benchmark models. The black line is the ground truth, the yellow line is the predicted value, and the red line is the start moment.}
\label{fig:lbsts}
\end{figure}

As illustrated in Figure \ref{fig:lbsts} and Table \ref{tab:lbs_per}, Apollo-Forecast significantly enhances accuracy in zero-shot scenarios, reducing WQL and MASE by up to 31.53\% and 14.29\%, respectively, compared to Chronos. These results demonstrate the robustness of the proposed method in pedestrian flow prediction, achieving substantial accuracy improvements without a significant increase in computational costs.

\subsubsection{Ablation Study}

\setlength{\tabcolsep}{1mm}

\begin{table}[ht]
    \centering
    \small
    \begin{tabular}{c|cccccc}
        \hline
        Horizon & \multicolumn{2}{c}{512} & \multicolumn{2}{c}{576} & \multicolumn{2}{c}{640} \\
        \hline
        Metric & WQL & MASE & WQL & MASE & WQL & MASE \\
        \hline
        Apollo-M & 5.98 & 17.15 & 4.41 & 17.11 & 6.23 & 18.07 \\
        w/o-RD & 5.82 & 16.83 & 4.35 & 17.01 & 6.19 & 17.98 \\
        w/o-AAQM & 9.31 & 18.53 & 6.05 & 20.50 & 9.40 & 21.61 \\
        Chronos-M & 9.34 & 18.46 & 6.84 & 20.61 & 9.47 & 21.71 \\
        \hline
    \end{tabular}
    \caption{Ablation analysis of Apollo-Forecast on the UCR dataset. WQL denotes aggregate relative WQL and MASE is aggregate relative MASE.}
    \label{tab:abl}
\end{table}

In our ablation study, we aimed to verify the impact of different components within Apollo-Forecast. Using the UCR dataset and setting the horizon range from 512 to 640, we conducted experiments with evaluation methods consistent with previous sections. We tested two variant implementations: w/o-RD, where the RD mechanism in the Apollo-Forecast framework was disabled, and w/o-AAQM, where AAQM was removed from the framework, thus no longer attenuating high-frequency noise.

Table \ref{tab:abl} shows the experimental results as follows:

\begin{enumerate}
    \item \textbf{RD}: Disabling RD (w/o-RD) slightly improves WQL but decreases MASE marginally.
    \item \textbf{AAQM}: Removing AAQM (w/o-AAQM) significantly worsens both WQL and MASE.
\end{enumerate}

In conclusion, both RD and AAQM are crucial for Apollo-Forecast's superior performance, with AAQM having a more significant impact on error reduction.

\subsection{Conclusion}

This study introduces the Apollo-Forecast approach, a novel framework addressing aliasing distortion and slow inference in TSFM. The Anti-Aliasing Quantization Module reduces distortion during tokenization, while Race Decoding accelerates inference. Extensive experiments on diverse datasets demonstrate its superiority over SOTA methods.

Future work will refine the framework and explore its applicability to other domains. We aim to enhance Apollo-Forecast's generalization and efficiency in multi-frequency scenarios, making it valuable for applications like finance, weather forecasting, and manufacturing.

\section{Acknowledgments}
This work was partly supported by the National Natural Science Foundation of China under grant 72374154 and 62273261.

\bigskip

\bibliography{aaai25}

\end{document}